\documentclass{article}
\usepackage{mystyle}

\usepackage{times}
\usepackage{xcolor}
\usepackage{soul}
\usepackage[utf8]{inputenc}
\usepackage[small]{caption}

\usepackage{booktabs} % For formal tables
\usepackage{dsfont}
\usepackage{stmaryrd}
\usepackage{color}
\usepackage{bm}
\usepackage{CJK}
\usepackage{algorithm}
\usepackage{algorithmicx}  
\usepackage{algpseudocode}  
\usepackage{amsmath}
\usepackage{indentfirst}
\usepackage{subfigure}
\usepackage{multirow}

\usepackage{diagbox}
\usepackage{array}
\usepackage{subfigure}
\usepackage{tikz}
\usepackage{epstopdf}
\usetikzlibrary{patterns}
\usepackage{footnote}
\usepackage{url}
\makesavenoteenv{tabular}
\makesavenoteenv{table}
\newcommand{\PreserveBackslash}[1]{\let\temp=\\#1\let\\=\temp}
\newcolumntype{C}[1]{>{\PreserveBackslash\centering}p{#1}}
\newcolumntype{R}[1]{>{\PreserveBackslash\raggedleft}p{#1}}
\newcolumntype{L}[1]{>{\PreserveBackslash\raggedright}p{#1}}

\title{MQGrad: Reinforcement Learning of Gradient Quantization in Parameter Server}
\author{
Guoxin Cui$^1$,
Jun Xu$^{1}$,
Wei Zeng$^1$,
Yanyan Lan$^{1}$,
Jiafeng Guo$^{1}$,
Xueqi Cheng$^{1}$,
\\
$^1$ CAS Key Laboratory of Network Data Science and Technology,\\
 Institute of ComputingTechnology, Chinese Academy of Sciences, Beijing 100190, China \\
}
\begin{document}

\maketitle

%\vspace{-3cm} 

\begin{abstract}
%Parameter server, which distributes the data and workload to worker nodes and maintains globally shared parameters on server nodes, 
%Parameter server (PS) has become a popular framework for solving large scale machine learning problems. 
One of the most significant bottleneck in training large scale machine learning models on parameter server (PS) is the communication overhead, because it needs to frequently exchange the model gradients between the workers and servers during the training iterations. Gradient quantization has been proposed as an effective approach to reducing the communication volume. One key issue in gradient quantization is setting the number of bits for quantizing the gradients. Small number of bits can significantly reduce the communication overhead while hurts the gradient accuracies, and vise versa. An ideal quantization method would dynamically balance the communication overhead and model accuracy, through adjusting the number bits according to the knowledge learned from the immediate past training iterations. Existing methods, however, quantize the gradients either with fixed number of bits, or with predefined heuristic rules. In this paper we propose a novel adaptive quantization method within the framework of reinforcement learning. The method, referred to as MQGrad, formalizes the selection of quantization bits as actions in a Markov decision process (MDP) where the MDP states records the information collected from the past optimization iterations (e.g., the sequence of the loss function values). During the training iterations of a machine learning algorithm, MQGrad continuously updates the MDP state according to the changes of the loss function. Based on the information, MDP learns to select the optimal actions (number of bits) to quantize the gradients. Experimental results based on a benchmark dataset showed that MQGrad can accelerate the learning of a large scale deep neural network while keeping its prediction accuracies. 
\end{abstract}

\section{Introduction}\label{sec:Intro}
%With the rapid growth of the training data and the resulting machine learning model complexity, distributed optimization and inference has become a popular solution for scaling up the machine learning problems. Using a cluster of computers can overcome the problem that no single machine can solve the problem efficiently, due to the huge volume of data, the intensive computational workloads, and the large number of model parameters. Parameters sever (PS)~\cite{Li:OSDI2014:PS} is one of the most popularly adopted distributed computing framework tailored for large scale machine learning. PS splits the computers in a cluster into worker nodes and server nodes. The data and workload are distributed to worker nodes and the globally shared model parameters are maintained by the server nodes. During the training of machine learning models, the worker nodes process data and calculate the gradients while server nodes synchronize parameters and perform global updates. 

With the rapid growth of the training data and the resulting machine learning model complexity, distributed optimization has become a popular solution for scaling up the machine learning problems. Parameters sever (PS)~\cite{li2014scaling} is one of the most popularly adopted distributed computing framework tailored for large scale machine learning. PS splits the computers in a cluster into worker nodes and server nodes. The data and workload are distributed to worker nodes and the globally shared model parameters are maintained by the server nodes. During the training of machine learning models, the worker nodes process data and calculate the gradients while server nodes synchronize parameters and perform global updates. PS can scale up a number of machine learning algorithms such as LDA and logistic regression.

It has been observed that the communication overhead is one of the major bottleneck in PS\cite{dean2012large,ho2013more}. At each of the optimization iteration, after finishing the local computations, multiple worker nodes need to push the resulting gradients of the parameters to the corresponding server nodes for parameter updating, and then pull the updated parameters to local for next iteration computations. Since the optimization procedure needs to execute a large number of iterations, the communication volume between the worker nodes and server nodes is huge and time consuming. Thus, how to reduce the communication volume becomes a key problem for accelerating the training of machine learning algorithms on PS.

Gradient quantization has been proposed as one of the most effective approach to reduce the communication overhead in distributed systems~\cite{oland2015reducing,seide20141,alistarh2016qsgd}. It reduces the number of bits used to transmit each parameter through quantizing (compressing) the transmitted values. When applying gradient quantization in PS, how to determine the number of bits used to transit each parameter, aka the quantization bits, is a critical issue. On one hand, one may want to set a small quantization bits for significantly reducing the communication overhead. On the other hand, the quantization bits cannot be too small because heavily compressing the gradients inevitably makes the model inaccurate, which may slow down the decreasing of the loss or even make the optimization not converge. How to balance between the communication overhead and gradient accuracy is one of the key issues in gradient quantization. 

Ideally, for choosing optimal quantization bits at each iteration,  PS system would dynamically adjust the bits with the knowledge learned from the past optimization iterations (e.g., the decreasing rates of the loss function at the past a few iterations). Existing approaches, however, usually choose a fixed  quantization bits before running the machine learning algorithm~\cite{seide20141,alistarh2016qsgd}. In recent years methods are proposed in which the quantization bits can be dynamically adjusted with predefined heuristic rules. For example, \citeauthor{oland2015reducing} proposes to adjust the quantization bits according to the norm of gradient vector. The experience from the past optimization iterations is not fully utilized.

\begin{figure}
\includegraphics[width=0.45\textwidth]{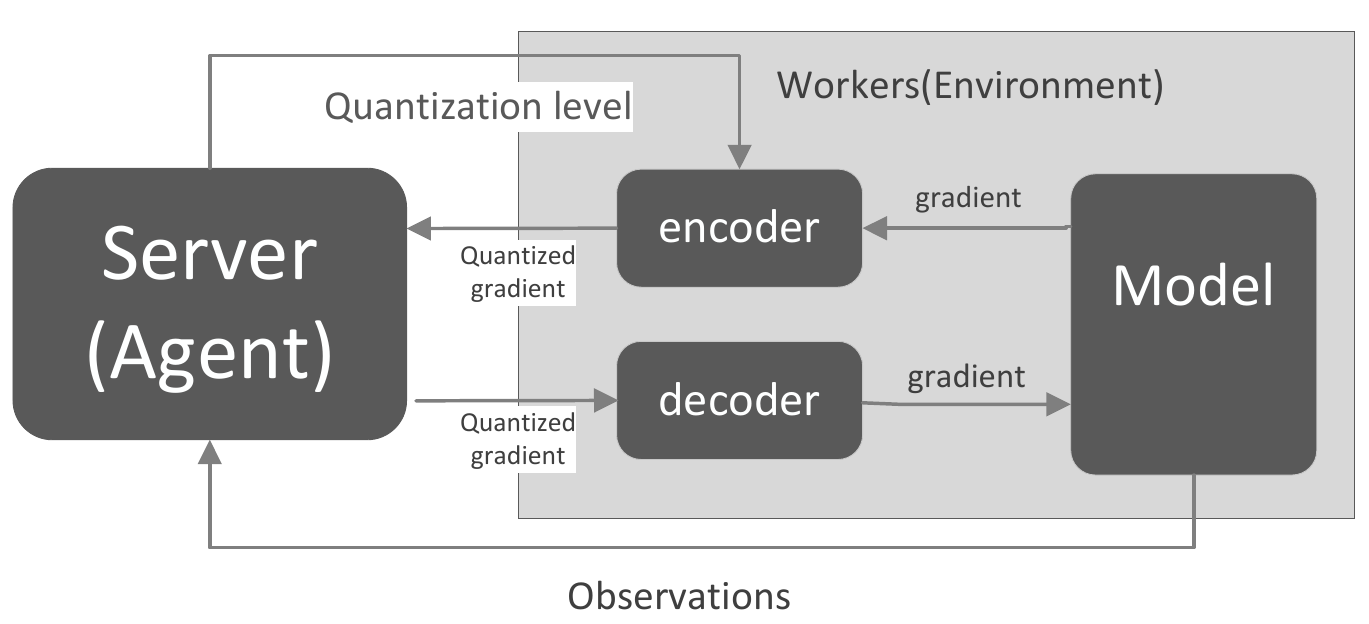}
  \caption{Agent-environment interaction in MQGrad.}\label{fig:simple_arch}
\end{figure}

In this paper, we aim at developing a new method that can learn to adjust the quantization bits, on the basis of the information collected from the past optimization iterations. Inspired by the success of learning to learn~\cite{andrychowicz2016learning} and reinforcement learning, we propose to use the Markov decision process (MDP) to learn the quantization bits in PS, referred to as MQGrad. The agent-environment interaction framework of MQGrad is shown in Figure~\ref{fig:simple_arch}. MQGrad formalizes the adjustment of the quantization bits as actions in an MDP. At each iteration of training the machine learning algorithm, the MQGrad agent repeatedly monitors the values of the machine learning loss function for updating its state and calculating the rewards. Then, it chooses the best action (quantization bits) and sends it to the workers for quantizing the gradients at the current iteration. The reinforcement algorithm of SARSA~\cite{sutton1998reinforcement} is utilized here for determining the quantization bits and updating the parameters of the MDP model.

MQGrad offers several advantages: ease in implementation, ability of balancing the communication overhead and model accuracy automatically, and effectively accelerating the large scale machine learning algorithms. 

Experimental results indicate that MQGrad can outperform the baseline quantization methods including the fixed quantization methods and the adaptive quantization, in terms of accelerating a deep learning algorithm trained on CIFAR-10 dataset.

\section{Related work}\label{sec:RelatedWork}
%\subsection{Reducing the communication overhead in distributed learning}\label{sec:RW:SRD}
A lot of research efforts have been spent for accelerating the distributed (machine learning) systems through reducing the communication cost of the system.  

% Reduce the number of parameters
One direct approach to minimize communication overhead is just to reduce the number of parameters need to be exchanged, e.g. by having fewer parameters in the first place by sparsifying them. In the early works\cite{lecun1990optimal,hanson1989comparing,hassibi1993second,strom1997phoneme}, network pruning has been proved as a valid way to reduce the complexity of the network. Recently, \citeauthor{han2016eie} pruned state-of-the-art CNN models with no loss of accuracy and \citeauthor{han2015deep} prunes the network's connections by removing all connections with weights below a threshold to reduce the parameters. 

% Reduce the number of bits
Another way to reduce the communication overhead is using less bits to represent the parameters or gradients, called parameter or gradient quantization. For example, \citeauthor{seide20141} uses 1 bit to quantize the gradients which greatly reduce the communication overhead while it needs the quantization error to be carried forward across mini-batches. \citeauthor{alistarh2016qsgd} proposed quantized SGD(QSGD) which is a family of compression schemes with convergence guarantees and good practical performance. \citeauthor{wen2017terngrad} used similar stochastic quantization like QSGD but focused on using three possible values to represent each value of  gradient. \citeauthor{oland2015reducing} reduced the communication overhead by nearly an order of magnitude through adaptively choosing the bits to quantize the weights according to gradient's mean square error. The method is based on the simple hypothesis: when the gradient's norm is large, more bits are needed to represent the gradient because relatively small perturbations can result in relatively large changes in the error.

%\subsection{Learning to learn}
All existing parameter quantization methods cannot utilize the knowledge from the optimization history. In this paper, we propose to learn to set the quantization bits, on the basis of the data collected from the past training iterations. The idea is similar to that of ``learning to learn''~\cite{andrychowicz2016learning} which automatically learns the updating rule of optimization in machine learning. A number of learning to learn algorithms has been proposed in the past a few years and reinforcement learning is also used for the task. For example, reinforcement learning algorithms are used for tuning the learning rate~\cite{fu2016deep,daniel2016learning,xu2017reinforcement}, for optimizing device placement for Tensorflow computational graphs\cite{mirhoseini2017device}, and for generating network architecture to maximize the expected accuracy of the validate set~\cite{zoph2016neural}. In this paper, we make use of the reinforcement learning model of MDP to learn the quantization bits for compressing the gradients.

\section{Our Approach: MQGrad}
In this section, we describe the proposed MQGrad model for reducing the communication overhead in PS. %, which aims at accelerating the distributed learning of machine learning models with gradient quantization guided by an MDP. 

\subsection{Background: machine learning with PS}

The goal of many machine learning algorithms can be formalized as minimizing a ``loss function'' which captures the properties of the learned model, e.g., the error in terms of the training data and the complexity of the learned model. In general, there is no closed-form solution for the optimization problem. Instead, the training starts from an initial model. It iteratively refines the model by processing the training data and stops when a (near) optimal solution is found or the model is considered to be converged. 
%Given a set of training data $\{(\mathbf{x}_i, y_i)\}_{i=1}^N$, where $\mathbf{x}_i$ and $y_i$ are the feature vector and the label for the $i$-th instance, respectively, and $N$ is the number of training instances, the machine learning algorithm typically minimizes the loss function to obtain the model:
% \[
% L(\mathbf{w})=\sum_{i=1}^N \ell(\mathbf{x}_i, y_i; \mathbf{w}) + \Omega(\mathbf{w}),
% \]
% where $\ell$ represents the prediction error on the training data and regularizer $\Omega$ penalize the model complexity. 

\begin{figure}[t]
\includegraphics[width=0.5\textwidth]{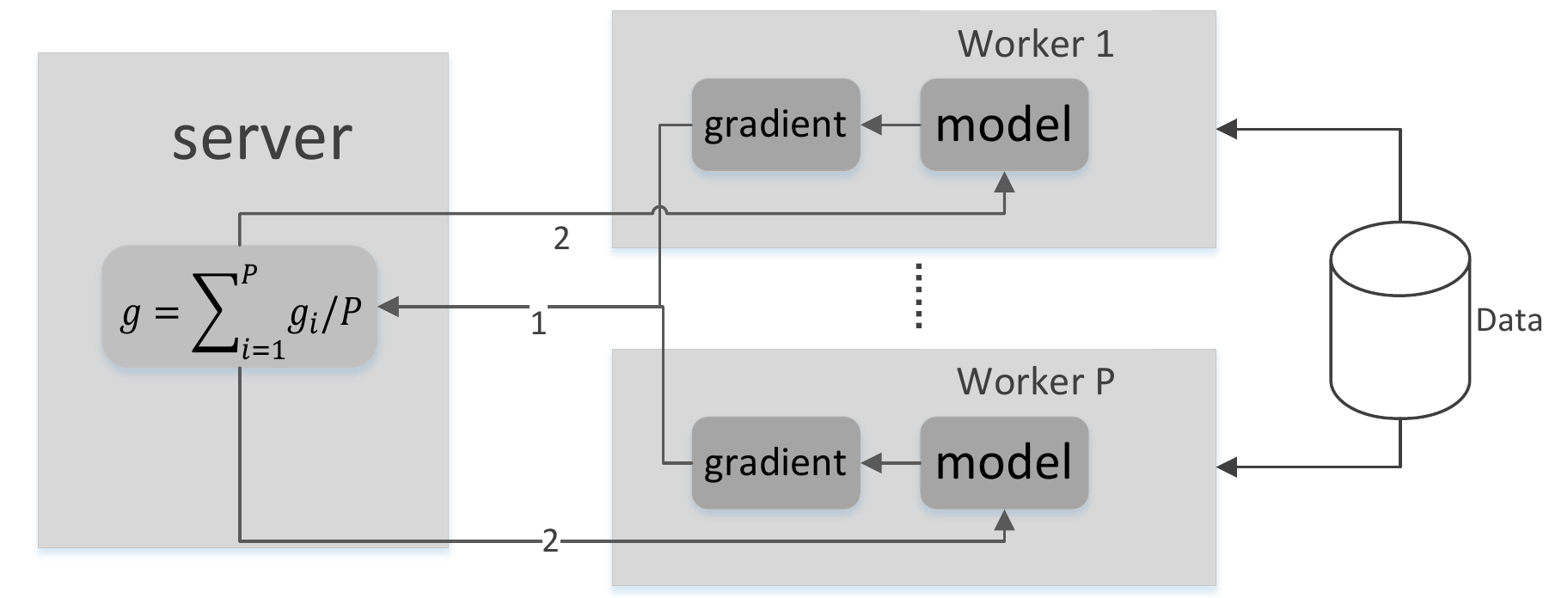}
\caption{Distributed training of machine learning algorithms in parameter server.}\label{fig:PS}
\end{figure}

In a distributed computing environment, we usually split the training data in multiple workers and send each worker the same parameters at first. Each worker computes gradient of the loss function with respect to parameters based on the training data it has and then sends the gradient to the server. Server collects gradients from all workers, averages it and then sends the gradient to all workers for updating the parameters. Figure~\ref{fig:PS} shows the process.

Usually a PS system spends a lot of time to communicate the gradients between sever nodes and worker nodes, as the number of parameters is large and they need to be transfered at each of the training iteration. Gradient quantization has been widely used to reduce the communication overhead and accelerate the training process. In the following sections, we describe our proposed MQGrad model which learns to adjust the quantization bits with reinforcement learning.

\subsection{MQGrad system architecture}
\begin{figure*}[hbt]
\begin{center}
\includegraphics[width=0.8\textwidth]{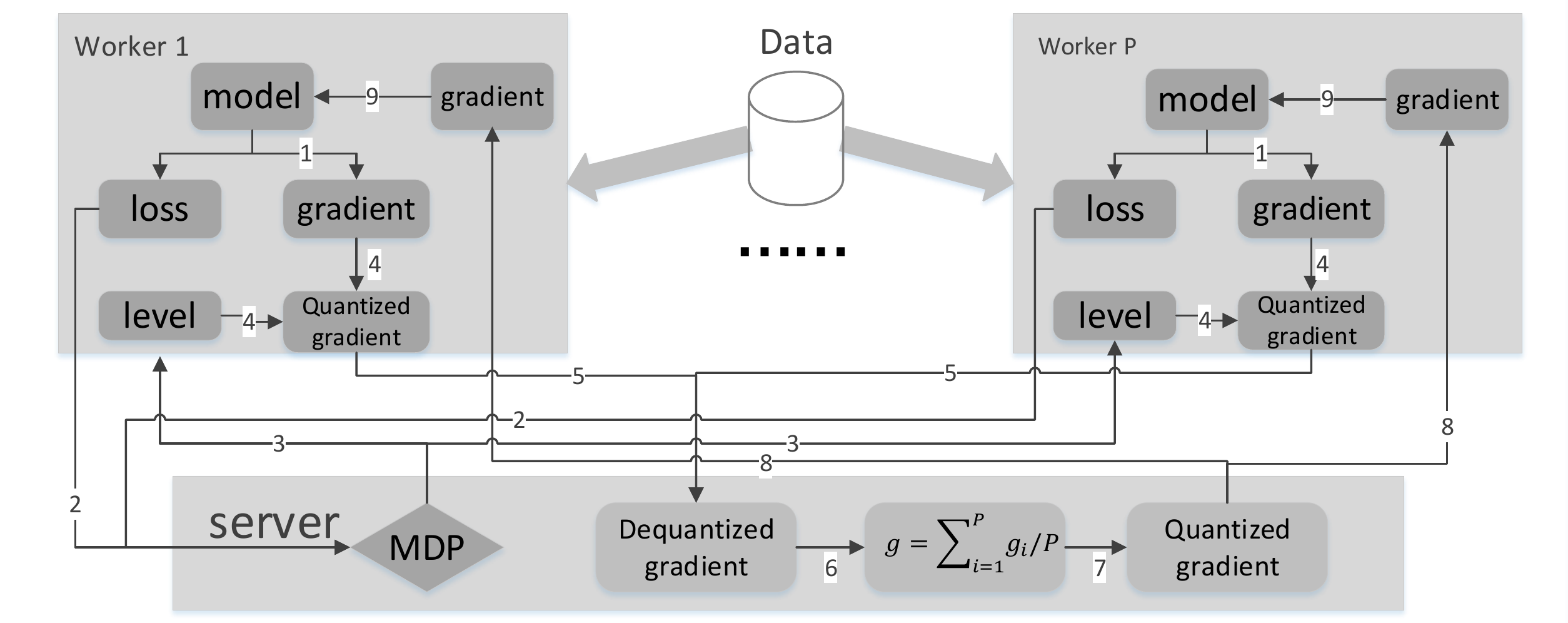}
\end{center}
\caption{MQGrad system architecture.}\label{fig:MQGradArch}
\end{figure*}

We extend the PS architecture shown in Figure~\ref{fig:PS} with an MDP module on the sever side and gradient quantize/de-quantize modules on all of the nodes, achieving the MQGrad system shown in Figure~\ref{fig:MQGradArch} and the functions executed on the scheduler, workers, and servers are shown in Algorithm~\ref{alg:MQGradFunctions}.

%Unlike existing method like fix bit quantization method and heuristic adaptive method, we use reinforcement learning to tune the quantization bits adaptively. Except the primary model to be optimized on every worker nodes, there is an reinforcement learning model on the server side which tells the quantization bits to use periodically.

%Further suppose that each worker contains a monitor that can observe the status of the algorithm being trained and collects the values of the loss function at each iteration. It is assumed that the losses are calculated on the basis of the local data at each worker. 

Suppose that a large scale machine learning model is being trained on the PS. After distributing the training data and the model parameters (necessary working set) to each worker node, the training algorithm executes an iterative optimization of its loss function. At each iteration $m$, given the current model parameters, the training algorithm calculates the local gradients at each worker node. At the same time, each worker also calculates the local value of the loss function based on the local data (step 1 in Fig.~\ref{fig:MQGradArch}, line 28 in Alg.~\ref{alg:MQGradFunctions}). The local values at all of the workers are collected by the sever MDP module (step 2 in Fig.~\ref{fig:MQGradArch}, line 38 of Alg.~\ref{alg:MQGradFunctions}). After that, the MDP module at server restores the overall global loss, updates its state, calculates the reward, determines the action (the quantization bits), and finally broadcasts the number to all worker nodes (step 3 in Fig.~\ref{fig:MQGradArch}, line 39-47 in Alg.~\ref{alg:MQGradFunctions}). Given the quantization bits, the worker nodes quantize\footnote{MQGrad uses the Quantize (Encode) and De-quantize (Decode) functions in \url{https://www.tensorflow.org/performance/quantization}.} their local gradients (step 4 in Fig.~\ref{fig:MQGradArch}, line 30 in Alg.~\ref{alg:MQGradFunctions}) and send the quantized local gradients to the parameter server (step 5 in Fig.~\ref{fig:MQGradArch}, line 31 in Alg.~\ref{alg:MQGradFunctions}). The server nodes de-quantize and summarize all of the received local gradients to a global gradient for updating the model parameters (step 6 and 7 in Fig.~\ref{fig:MQGradArch}, line 51-52 in Alg.~\ref{alg:MQGradFunctions}). Then the server broadcasts the quantized global gradient (step 8 in Fig.~\ref{fig:MQGradArch}, line 53 in Alg.~\ref{alg:MQGradFunctions}) and the workers receive it, de-quantize the gradient, and update the local model parameters (step 9 in Fig.~\ref{fig:MQGradArch}, line 32-33 in Alg.~\ref{alg:MQGradFunctions}). 

%Note that  in Algorithm~\ref{alg:MQGradFunctions} use method in . Function Encode quantizes a gradient to a quantized gradient and function Decode restores a gradient from a quantized gradient.
%Similar to that of used in conventional PS systems, the global gradient is used for updating the model parameters on the server as well as the model parameters on worker nodes. Thus, the global gradient is quantized with the same quantization bits as just used at worker nodes. 

Receiving the signal that the model parameters have been updated, the machine learning training algorithm moves to iteration $m+1$ and re-estimates the local gradients and local losses. The process repeats until converge or the number of iterations reaches a predefined maximum number.

%the MQGrad monitors the process and builds an MDP for controlling the quantization bits for transiting the gradients in  Line 12.

%The architecture of MQGrad is shown in Figure~\ref{fig:MQGradArch}. The whole procedure is driven by the worker nodes and reinforcement learning is executed on the servers. Each worker node keeps a quantization bits for quantizing the gradient vectors. At each of the learning iteration, all of the worker nodes compute the local gradients and the local losses, and send the losses to the server nodes for constructing the overall MDP state. When the server nodes have received all workers' losses, it must decide if it is time to use the MDP to change the quantization bits. In our algorithm, we choose to use MDP to tune quantization bits every $m$ iterations. If it's not the time to tune quantization bits, server just sends the same quantization bits as the last iteration. If it is time to tune quantization bits, server uses MDP to get a new quantization bits and then send it to all workers. The worker nodes quantize the gradients using the received quantization bits and send quantized gradient to the server. Server then receives all worker's quantized gradient, dequantizes them, averages them, quantize the averaged gradient using current quantization bits and at last sends the quantized gradient to all workers. When the worker receives the quantized gradient, it dequantizes it and uses it to update parameters.

% 1. Workers compute local gradients and losses. 2. Workers send losses to MDP module. 3. Server sends the number of bits to all workers. 4. Workers quantize the gradients. 5. Workers send quantized gradients to server. 6. Server de-quantizes local gradients and aggregate to global gradient. 7. Sever quantize the global gradient. 8. Server sends the quantized global gradient to workers. 9. Workers de-quantize the global gradient and update model parameters.}

 \begin{algorithm}[htp]
\caption{MQGrad functions}\label{alg:MQGradFunctions}
\renewcommand{\algorithmicrequire}{\textbf{Input:}}
\renewcommand{\algorithmicensure}{\textbf{Output:}}
\label{alg:workerAndServerFunc}
\begin{algorithmic}[1]

\State{\textbf{\underline{Task scheduler:}}}
\State{issue \Call{LoadData}{ } to all workers}
\State{init $\mathbf{w}$ and issue it go all workers}
\For{iteration $m=0$ to $M$}
	\State{Issue \Call{WorkerIterate}{$p,m$}, where $p$ is worker ID}
\EndFor

\State{}

%following are workers' function
\State{\textbf{\underline{Worker $p=1,\cdots,P$}:}}
\Function{LoadData}{ } 
	\State{load a batch of training data $\{\mathbf{x}_{ip}, y_{ip}\}_{i=1}^{N_p}$}
\EndFunction

\State{}

\Function{SendLossThenReceivebits}{$L_{p, m}$}
	\State{send $L_{p, m}$ to server}
    \State{remote call server function \Call{ReceiveLossThenSendbits}{$L_{p, m}$}}
    \State{receive quantize bits $K$ from server}
    \State{\Return{$K$}} 
\EndFunction

\State{}

\Function{SendQGThenReceiveQG}{$\mathbf{\bar{g}}_{p, m}$}
	\State{send $\mathbf{\bar{g}}_{p, m}$ to server}
    \State{remote call server function \Call{ReceiveQGThenSendQG}{$\mathbf{\bar{g}}_{p, m}$}}
    \State{receive updated gradient $\mathbf{\bar{g}}^m$ from server}
    \State{\Return{$\mathbf{\bar{g}}^m$}}
\EndFunction

\State{}

\Function{WorkerIterate}{$p, m$}
	\State{$(\mathbf{g}_{p, m}, L_{p, m})\leftarrow$ Gradient/loss w.r.t. a batch of data}
    \State{$K \leftarrow$ \Call{SendLossThenReceivebits}{$L_{p, m}$}}
    \State{$\mathbf{\bar{g}}_{p, m} \leftarrow$ \Call{Quantize}{$\mathbf{g}_{p, m}, K$}}
    \State{$\mathbf{\bar{g}}^m \leftarrow$ \Call{SendQGThenReceiveQG}{$\mathbf{\bar{g}}_{p, m}$}}
    \State{$\mathbf{g}^m \leftarrow$ \Call{De-quantize}{$\mathbf{\bar{g}}^m$}}
    \State{$\mathbf{w} \leftarrow \mathbf{w} - \eta \mathbf{g}^m$}
\EndFunction

\State{}

% following are server's function
\State{\underline{\textbf{Servers:}}}

\Function{ReceiveLossThenSendbits}{$L_{p, m}$}
	\State{receive loss $L_{p, m}$'s from workers}
    \If{all $L_{p, m}, p =1, \cdots, P$ are received}
    	\State{$L^m \leftarrow \frac{\sum_{p=1}^P L_{p, m}}{P}$}
    	\If{$m\%T = 0$}
    		\State{$t \leftarrow m/T$}
    		\State{$K \leftarrow$ \Call{MQGrad-SARSA}{$t$}}
            \State{send $K$ to all workers} 
    	\EndIf
      \EndIf
      \State{send the last iteration $K$ to workers}
\EndFunction

\State{}

\Function{ReceiveQGThenSendQG}{$\mathbf{\bar{g}}_{p, m}$}
	\State{$\mathbf{g}_{p, m} \leftarrow$ \Call{De-quantize}{$\mathbf{\bar{g}}_{p, m}$}}
    \If{all $\mathbf{g}_{p, m} , p =1, \cdots, P$ are received}
		\State{$\mathbf{g}^m\leftarrow \frac{\sum_{p=1}^P \mathbf{g}^{p, m}}{P}$}
		\State{$\mathbf{\bar{g}}^m \leftarrow$ \Call{Quantize}{$\mathbf{g}^m$, $K$}}
		\State{send $ \mathbf{\bar{g}}^m $ to all workers}
     \EndIf
\EndFunction

\end{algorithmic}
\end{algorithm}

 \begin{algorithm}[hbt]
\caption{MQGrad-SARSA}\label{alg:MQGradSARSA}
\renewcommand{\algorithmicrequire}{\textbf{Input:}}
\renewcommand{\algorithmicensure}{\textbf{Output:}}
\begin{algorithmic}[1]
\Require{MDP time step $t$}
%\Ensure{$K$}
% \Function{ConstructState}{$\textbf{s}$}
% 	\State{$\textbf{s}'.\overline{L} \gets $}
% 	\State{check $s.bits$ and $s.bits+1$ 's q values to get the $s_{new}.bits$ }
%     \State{add loss information to $s_{new}$}
%     \State{\Return $s_{new}$} 
% \EndFunction

% \State{}

% \Function{GetReward}{ }
% 	\State{$r \leftarrow $get reward from queue}
%     \State{\Return{$r$}} 
% \EndFunction

% \State{}

% \Function{Getbits}{$\textbf{s}, a$}
% 	\If{$a = a_1$}
%     	\State{\Return{$\textbf{s}.bits + 1$}}
%         \Else \If{$a = a_0$}
%         	\State{ \Return{$\textbf{s}.bits$}}
%         \EndIf
%     \EndIf
% \EndFunction

% \State{}

%\Function{MDP-SARSA} {$\overline{\mathbf{L}}^t, c_t$}
	%\State{push $L^j$ to $queue$}
	%\If{$m$ devides $j$}
    	\State{${s}_t \leftarrow T({s}_{t-1}, a_{t-1}$)}
        \State{$r_{t-1} \leftarrow R({s}_{t-1}, a_{t-1})$ }
        \State{Choose action $a_t$ from ${s}_t$ using policy derived from $Q$}
		\State{$\textbf{v}\leftarrow \textbf{v} + \eta[r_{t-1} + \gamma Q({s}_t,a_t)-Q({s}_{t-1},a_{t-1})]\frac{\partial Q}{\partial \textbf{v}}|_{t-1}$}	
		%\State{$\textbf{s}\leftarrow \textbf{s}'; a\leftarrow a'$}
        \State{\Return{$\textbf{s}_t.n_t + a_t$}};
        %\State{clean $queue$}
    %\EndIf
    %\State{\Return{$K$}}
%\EndFunction
\end{algorithmic}
\end{algorithm}

\subsection{Learning for gradient quantization with MDP}
The key component in MQGrad is the MDP module which determines the quantization bits. The configuration of the MDP is as follows: 

\textbf{Time step $t$}: $t\in Z^+ \cup \{0\}$ is the discrete time step of the MDP. To avoid adjusting the quantization bits too frequently, which may result in an unstable training process, the MDP model in MQGrad is configured to update the quantization bits every $T$ training iterations. That is, the server will broadcast the identical quantization bits used in the last iteration to worker nodes (step 3 in Figure~\ref{fig:MQGradArch}) if $m\%T \neq 0$, where $m$ is the iteration number of the machine learning training algorithm. During these iterations, the MDP module only collects the losses for constructing its state. The MDP module will be activated to update the quantization bits when $m\%T = 0$. Thus the MDP time step $t = \lfloor \frac{m}{T}\rfloor$. In this paper, we empirically set $T=5$, which means the MDP time step is 5 times slower than the number of training iterations. Note that both $t$ and $m$ start from 0.

\textbf{States $\mathcal{S}$}: The MDP state $\textbf{s} \in \mathcal{S}$ at time step $t$ is denoted as $\textbf{s}_t =\left[n_t, \overline{\mathbf{L}}_t\right]$, where $n_t \in Z^+$ is the most confident quantization bits at time step $t$, predicted by the $Q$ function; $\overline{\mathbf{L}}_t$ is calculated as follows: at time step $t$, the MDP receives $T$ consequent global losses, denoted as $\mathbf{L}_t=\{L_t^1, \cdots, L_t^{T}\}$, where $L_t^{i} (i=1\cdots T)$ is the global loss calculated based on the local losses received at the training iteration $(t-1)\times T + i$. The $T$ global losses reflect the goodness of the quantization bits used at the last $T$ iterations. These values, however, may vary to a large extent. To make the statistics of these losses stable, following the practices in~\cite{daniel2016learning}, MQGrad makes use of the moving average technique to smooth these losses: 
\begin{equation}\label{eq:movingAvg}
\overline{L}_t^i = \left\{
\begin{array}{rcl}
\alpha * L_t^i + (1 - \alpha)*\overline{L}_t^{i-1} &  & 1 < i \leq T \\
\alpha * L_t^1 + (1 - \alpha) * \overline{L}_{t-1}^{T} &  &  i = 1,
\end{array} \right. 
\end{equation} 
where $\alpha$ is the parameter. Thus $\overline{\mathbf{L}}$ is a sequence of $T$ values: $\overline{\mathbf{L}}_t= \{\overline{L}_t^1, \cdots, \overline{L}_t^{T}\}$.

%We design the state at time step $t$ as a tuple $\mathbf{s_t} = \{\mathbf{\overline{L_t}}, bits_t\}$. $bits_t$ is an integer that means the current bits using to quantize gradient.

\textbf{Actions $\mathcal{A}$}: MQGrad has two actions at each time step: $\mathcal{A} = \{0, 1\}$, where $0$ keeps the current quantization bits and $1$ increases the quantization bits by one. Thus given $\textbf{s}_t$ and the chosen action $a_t$, the quantization bits for the immediate next $T$ training iterations is $n_t + a_t$, where $n_t$ is the quantization bits in state $\textbf{s}_t$. Note that MQGrad does not decrease the quantization bits. The configuration is based on the observation that with the machine learning training iteration goes on, more accurate gradients are needed to update the model parameters because the parameters are closer to the optimal solution. Experimental results also showed that the configuration can achieve better results.

\textbf{$Q$ function}: The $Q$ function predicts the value of taking action $a$ at the state $\textbf{s}$ following policy $\bm\pi$. Following the practice in DQN~\cite{mnih2013playing}, MQGRad configures the $Q$ function as a neural network (parameterized by $\textbf{v}$). The input to the neural network is the state and the outputs are the confidence values for the available actions. The parameter $\textbf{v}$ will be updated during the MDP iterations with SARSA algorithm~\cite{sutton1998reinforcement}.

\textbf{Policy $\bm\pi$}:$\bm\pi$ defines the probability of selecting an action $a$ at state $\textbf{s}$. We define the policy $\bm\pi$ with the $\epsilon$-greedy criteria for balancing the exploration and exploitation. Specifically, at the MDP time step $t$, given the state $\textbf{s}_t$, the probability of selecting an action $a_t$ is denoted as $\pi(a_t|\textbf{s}_t)$ and defined as:
$$ 
\pi(a_t|\textbf{s}_t) = \left\{
\begin{array}{rcl}
1 - \epsilon &  &  a_t = \mathop{\arg \max}_a  Q(\textbf{s}_t, a)\\
\epsilon &  &  \mathrm{otherwise},
\end{array} \right. 
$$

\textbf{Reward $R$}: MQGrad calculates the reward on the basis of the $T$ consequent losses collected from the last $T$ training iterations and the total time cost for executing the $T$ iterations. Intuitively, small decrease in loss with high time cost makes the reward small, and vise versa. Specifically, suppose that the moving averaged losses are $\overline{\mathbf{L}}_{t+1} = \{\overline{L}_{t+1}^1, \cdots, \overline{L}_{t+1}^{T}\}$ and the time cost for executing the last $T$ training iterations is $c_{t+1}\in Z^+$ (in milliseconds). MQGrad solves the following linear regression problem for getting the decreasing rate with respect to iteration $\beta$:
\[
\begin{split}
(\beta, b)\leftarrow \arg\min_{\beta, b} \sum_{i=1}^{T}\left(\beta \times i + b - \overline{L}_{t+1}^i\right)^2,
\end{split}
\]
where $b$ is the bias. The reward is calculated as:
\begin{equation}
\begin{split}
R({s_t}, a_t) &= -\frac{1}{c_{t+1}}\times \beta \times \gamma,
\end{split}
\end{equation}
where $\gamma> 0$ is a scaling parameter. 

\textbf{Transition $T$}: The transition function $T:\mathcal{S}\times\mathcal{A}\rightarrow \mathcal{S}$ defines the transition of the MDP state. The output of $T$ also consists of two parts: $\textbf{s}_{t} = [n_{t}, \mathbf{\overline{L}}_{t}]$. These two components are calculated as: 
\begin{equation}
\begin{split}
\textbf{s}_{t} = [n_{t}, \mathbf{\overline{L}}_{t}] =& T(\mathbf{s}_{t-1} = [n_{t-1}, \mathbf{\overline{L}}_{t-1}], a_{t-1})\\
 = & [n_{t-1} + \arg\max_{a\in\{0, 1\}} Q(\textbf{s}_{t-1}, a), \mathbf{\overline{L}}_{t}].
\end{split}
\end{equation}
After selecting $a_{t-1}$ on the basis of ${s}_{t-1}$, the server broadcasts the quantization bits $n_{t-1} + a_{t-1}$ (where $n_{t-1}$ is the quantization bits in ${s}_{t-1}$) to all of the worker nodes. MQGrad then monitors and collects the losses of the immediate $T$ training iterations and constructs a sequence of moving averaged losses $\mathbf{\overline{L}}_{t}=\{\overline{L}_{t}^1, \cdots, \overline{L}_{t}^{T}\}$, on the basis of Equation~(\ref{eq:movingAvg}). As for $n_{t}$, if $Q({s}_{t-1}, 0) \geq Q({s}_{t-1}, 1)$, MQGrad keeps $n_{t} = n_{t-1}$. Otherwise, MQGrad increases the quantization bits in state ${s}_{t}$ by 1.

% Then we check that if $Q(\mathbf{s_t}, a_0) > Q(\mathbf{s_t}, a_1)$. If it is true, then we set $bits_{t+1}$ to $bits_{t}$. If not, we set $bits_{t+1}$ to $bits_{t} + 1$. We can conclude it as follows:
% \begin{equation}
% \begin{split}
% \mathbf{s_{t+1}} &= T(\mathbf{s_t}, a_t)\\
% \end{split}
% \end{equation}
% $$ = \left\{
% \begin{array}{rcl}
% \{\mathbf{\overline{L_{t+1}}}, bits_t\}  &  &  Q(\mathbf{s_t}, a_0) > Q(\mathbf{s_t}, a_1)\\
% \{\mathbf{\overline{L_{t+1}}}, bits_{t}+1\} &  &  Q(\mathbf{s_t}, a_0) \leq Q(\mathbf{s_t}, a_1)
% \end{array} \right. $$

% $$\mathbf{s_{t+1}} = T(\mathbf{s_t}, a_t) = \left\{
% \begin{array}{rcl}
% \{\mathbf{loss_{t+1}}, bits_t\}  &  &  Q(\mathbf{s_t}, a_0) > Q(\mathbf{s_t}, a_1)\\
% \{\mathbf{loss_{t+1}}, bits_{t}+1\} &  &  Q(\mathbf{s_t}, a_0) \leq Q(\mathbf{s_t}, a_1)
% \end{array} \right. $$

% \begin{figure}[hbt]
% \begin{center}
% \includegraphics[width=0.4\textwidth]{figure/sarsa-show.png}
% \end{center}
% \caption{The occasion of SARSA to run.}\label{fig:sarsa-show}
% \end{figure}
During the running of the MDP, the SARSA algorithm is used for determining the quantization bits and learning the parameters in $Q$ function, as shown in Algorithm~\ref{alg:MQGradSARSA}. The running of the MDP in MQGrad can be described as follows: at each MDP time step $t = 0, 1, \cdots $, the agent(server) receives the state ${s}_t=[n_t, \overline{\mathbf{L}}_t]$ (line 1 of Alg.~\ref{alg:MQGradSARSA}) and the reward $r_{t-1}$ (line 2 of Alg.~\ref{alg:MQGradSARSA}). Then an action $a_t$ is selected on the basis of the policy $\pi(a_t|\textbf{s}_t)$ (line 3 of Alg.~\ref{alg:MQGradSARSA}). After that, the system updates the parameter $\textbf{v}$ of the $Q$ network (line 4 of Alg.~\ref{alg:MQGradSARSA}). Finally the number of bits ${s}_t.n_t + a_t$ is returned for conducting the gradient quantization (line 5 of Alg.~\ref{alg:MQGradSARSA}).

The source code of MQGrad can be found in the Github \url{http://hide_due_to_annonymous_review}.

\section{Experiments}
\subsection{Experimental settings}
To test the performances of the proposed MQGrad system, experiments were conducted on two PS clusters. One consists of 12 nodes and the other consists of 18 nodes. Each nodes in the clusters contains 4 cores each of which has a frequency of 2.3GHz and these nodes were connected by a network with ~10MB/s bandwidth. 

The experiments were conducted on the basis of the CIFAR-10~\cite{krizhevsky2009learning} dataset. The machine learning algorithm tested is a 5-layer neural network: the first two are convolutional layers with each layers' parameter's shape being $[5,5,3,64]$ and $[5,5,64,64]$. Local response normalization after max-pooling is used~\cite{krizhevsky2012imagenet}. The third and fourth layers are fully connected layers with shapes $[3136, 2304]$ and $[2304, 3840]$, respectively. The last softmax layer is also a fully connected layer with shape $[3840, 10]$. Cross entropy loss with  $\ell_2$ norm of the third layer and fourth layer's parameters are used as the loss function. During the training, the batch size is set to 32 and the learning rate is set to 0.2. Considering the third and fourth layers have about 99.2\% of the network parameters, gradient quantization is applied to these two layers. Other parameters are communicated without any compression.

The range of quantization bits is set to 2 to 8 bits (7 levels). The $Q$ function has three layers: the first layer contains 5 nodes, representing the 5 average smoothed values in state $s$. The second layer contains 10 nodes with ReLU activation. The third layer is a linear layer which has 7 nodes, each corresponds to a quantization bits. %For example, if the bits $n$ in $\textbf{s}$ is 3 and the action $a$ to be chosen is $a_0$, $Q(\textbf{s}, a)$ is represented by the second value of the output layer of the network. 

MQGrad has some hyper parameters. The variables in SARSA $\epsilon =0.1$ and $\eta = 0.1$. The moving average parameter $\alpha = 0.01$ and the scaling variable $\gamma=300$.

We compared MQGrad with several state-of-the-arts baselines in gradient quantization, including the adaptive quantization method~\cite{oland2015reducing} (denoted as ``Adaptive'' in the paper) and the fixed quantization methods. For the fixed bit quantization methods, the numbers of quantization bits were set to 2, 4, and 8 and denoted as ``Fix (2-bit)'', ``Fix (4-bit)'', and ``Fix (8-bit)'', respectively.%In ``Adaptive'' method, the number of quantization bits is set to $2 + \frac{Z}{0.0005}$ where $Z$ is norm of the gradient to be quantized, averaged over all of the worker nodes.  

\subsection{Experimental results}
\begin{figure}[t]
\includegraphics[width=0.5\textwidth]{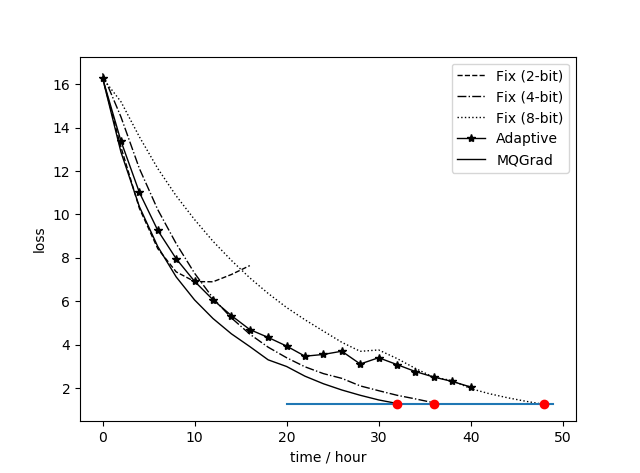}
  \caption{Learning curve on the 12-node cluster.}\label{fig:12_worker}
\end{figure}

\begin{figure}[t]
\includegraphics[width=0.5\textwidth]{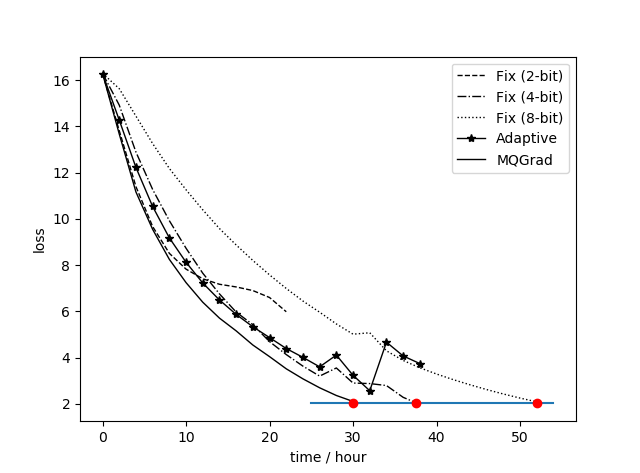}
  \caption{Learning curve on the 18-node cluster.}\label{fig:18_worker}
\end{figure}

%result(adaptive baseline detail)
Figure~\ref{fig:12_worker} and Figure~\ref{fig:18_worker} show the training curves of ``MQGrad'' as well as the baselines in terms of the neural network loss being optimized, on the 12-node PS cluster and the 18-node PS cluster, respectively. The x-axises indicate the training time (in terms of hours). From the results, we can see that ``MQGrad'' outperformed all the baseline methods (used less training time to reach smaller loss) on both of these two clusters. For example, compared with the best baseline ``Fix (4-bit)'', ``MQGRad'' used less 7.5 hours to reach the same loss on the 18-node cluster. The results indicate the effectiveness of using reinforcement learning for gradient quantization.  
%To make the curve smooth, we used moving average to losses of each iteration like that introduced in Section~\ref{state} except the moving average parameter was set to 0.02.

From the results, we can also see that the training curve ``Fix (2-bit)'' decreased the loss function very fast during the first ten hours. However, it did not converge in the remaining training time. The phenomenon indicated that at the early stage of the training low quantization bits helped to minimize the loss function fast. However, with the training goes on, high accurate gradients were necessary and the low quantization bits hurt the convergence. On the other hand, the training curve of ``Fix (8-bit)'', which used more bits for quantizing the gradients during the training, steadily decreased during all of the training time. However, the decreasing speed was slow because a lot of time was wasted for transiting the gradients. Thus, ``Fix (8-bit)'' needed longer time to converge. ``MQGrad'' made a good trade-off: it used low quantization bits at the early training stages for saving the communication volume, and gradually increased the quantization bits for increasing the gradient accuracies. The method of ``Adaptive'' can also decrease the communication volume at the early training stages. However, the predefined heuristics in ``Adaptive'' cannot make good decisions to guarantee the gradient accuracy at the later training phases.     
%Experiment results show that our algorithm outperforms all baselines. Figure \ref{fig:12_worker} shows that our method runs 4 hours faster than the best baseline when the loss reaches a low value. 

%result(accuracy)
% 12 workers

We also tested the model accuracies for these methods. Table~\ref{table:accuracy_12_worker} and Table~\ref{table:accuracy_18_worker} show the results on the 12-node cluster and 18-node cluster, respectively. ``N/A'' indicates the result is not available because the model has converged at the time. From the results we can see that the accuracies of MQGrad are higher than the baselines when trained with the same time, indicating the lower loss leads to higher performances. The final converged performances of MQGrad are comparable to ``Adaptive'' and ``Fix (8-bit)'', indicating MQGrad can accelerate the training process while keeping model accuracies. 

\begin{table}[t!]
\caption{Test accuracies (\%) of models trained on 12-node cluster.} \label{table:accuracy_12_worker}
\centering
\footnotesize
\begin{tabular}{c|cccc}
\hline 
\diagbox{method}{time} & 5 hours & 15 hours & 30 hours & 40 hours \\
\hline\hline Fix (2-bit) & 50.8 & 55.9 & N/A & N/A  \\
 Fix (4-bit) & 41.9 & 57.5 & 66.7 & N/A \\
 Fix (8-bit) & 35.7 & 55.6 & 60.8 & 68.1\\
 Adaptive & 49.4 & 60.1 & 65.3 & N/A \\
 MQGrad & 54.7 & 65.6 & 67.7 & N/A \\
\hline
\end{tabular}
\end{table}

\begin{table}[t!]
\caption{Test accuracies (\%) of models trained on 18-node cluster.} \label{table:accuracy_18_worker}
\centering
\footnotesize
\begin{tabular}{c|cccc}
\hline 
\diagbox{method}{time} & 5 hours & 15 hours & 30 hours & 40 hours\\
\hline\hline Fix (2-bit) & 51.5 & 58.5 & N/A & N/A  \\
 Fix (4-bit) & 45.4 & 57.9 & 68.4 & N/A\\
 Fix (8-bit) & 42.1 & 58.2 & 65.9 & 68.2\\
 Adaptive & 43.6 & 57.3 & 63.4 & N/A \\
MQGrad & 51.5 & 62.0 & 68.2 & N/A \\
\hline
\end{tabular}
\end{table}

\begin{table}[t!]
\newcommand{\tabincell}[2]{\begin{tabular}{@{}#1@{}}#2\end{tabular}}
\caption{Fraction of time cost (\%) caused by gradient quantization. } \label{table:AddtionTimeCost}
\centering
\footnotesize
\begin{tabular}{c|ccccc}
\hline 
 & \tabincell{c}{Fixed\\ (2-bit)}  & \tabincell{c}{Fixed\\ (4-bit)}  & \tabincell{c}{Fixed\\ (8-bit)}  & Adaptive & MQGrad\\
\hline\hline
12-node cluster & 13.3 & 8.0 & 4.44 & 8.13 & 7.41  \\
18-node cluster & 11.4 & 7.27 & 4.21 & 5.84 & 4.11 \\
\hline
\end{tabular}
\end{table}

Note that the execution of the quantization/de-quantization (and the MDP) modules in the baselines and MQGrad needs some additional time. We conducted experiments to show the fraction of these additional time among the whole training time. From the results shown in Table~\ref{table:AddtionTimeCost}, we can see that on the 12-node and the 18-node clusters, MQGrad respectively need 7.41\% and 4.11\% of the time for running the quantization, de-quantization, and the MDP modules. For other baseline methods, most of the fractions are less than 10\%. The results indicate that 1) the additional time costs caused by MDP module in MQGrad is negligible; 2) the time cost for quantizing/de-quantizing gradients is not high, making all of these methods can accelerate the overall training iterations.

\section{Conclusion}
In the paper we propose a novel gradient quantization method called MQGrad, for accelerating the distributed machine learning algorithms on parameter server. MQGrad learns to determine the number of bits for gradient quantization with the information collected from the past optimization iterations. MDP is used to formalize the process and the on-policy learning algorithm SARSA is used to learn the quantization bits and update the MDP parameters. Experimental results on a benchmark dataset showed that MQGrad outperformed the state-of-the-arts gradient quantization methods, in terms of accelerate the speeds of learning large scale machine learning models. Analysis showed that MQGrad accelerated the learning speeds through lowering the communication volume at the early stage of training and gradually improving the gradient accuracies with the training went on.

% initial runs of your .tex file t
%\bibliographystyle{named}
%\bibliography{ijcai18}  % sigproc.bib is the name of the Bibliography in this case

% ACM needs 'a single self-contained file'!
%\balancecolumns
%\balancecolumns % GM June 2007
% That's all folks!
\end{document}